\documentclass[10pt,twocolumn,letterpaper]{article}

\usepackage[pagenumbers]{cvpr} 

\usepackage[dvipsnames]{xcolor}
\newcommand{\red}[1]{{\color{red}#1}}
\newcommand{\blue}[1]{{\color{blue}#1}}

\usepackage{tikz}
\usepackage{pgfplots, pgfplotstable}
\pgfplotsset{compat=1.17}
\usetikzlibrary{colorbrewer} 

\usepackage{graphicx}
\usepackage{amsmath}
\usepackage{amssymb}
\usepackage{booktabs}

\usepackage{comment}
\usepackage{color}

\usepackage{caption}
\usepackage{boldline}
\usepackage{colortbl}
\usepackage{booktabs}
\usepackage{bbm}

\usepackage{multirow}
\usepackage{multicol}
\usepackage{makecell}
\usepackage{xcolor}
\usepackage{xspace}
\usepackage{float}

\usepackage{xifthen} 

\usepackage[accsupp]{axessibility}

\newcommand{\alg}{POLED\xspace}
\newcommand{\algpoi}{\alg-Poi\xspace}
\newcommand{\minisection}[1]{\paragraph*{\textbf{#1}}}
\newcommand{\ON}{\blue{ON}}
\newcommand{\OFF}{\red{OFF}}

\newcommand{\ek}{$e_{k}$}

\newcommand{\evSO}{\mathcal{E}} 
\newcommand{\evSD}{\mathcal{E}^{\alpha}} 
\newcommand{\fDown}{\Phi_{\alpha}} 
\newcommand{\pa}{\text{P}_{\evSO}(\mathbf{x}_k)} 
\newcommand{\evSOTnT}[2]{$\evSO_{{#1} \rightarrow {#2}}$}

\newcommand{\evSOTnTneq}[2]{\evSO_{{#1} \rightarrow {#2}}}
\newcommand{\evSDTnTneq}[2]{\evSD_{{#1} \rightarrow {#2}}}

\newcommand{\pdf}{$f(\cdot)$}
\newcommand{\pdfSO}{$f_{\evSO}$}
\newcommand{\pdfSOneq}{f_{\evSO}}
\newcommand{\pdfP}{f_{p}}
\newcommand{\pdfSOTnT}[2]{$f(\evSO_{{#1} \rightarrow {#2}})$}

\newcommand{\score}{\mathcal{S}}

\newcommand{\best}[1]{\textbf{#1}}

\newcommand{\pmsize}[1]{\footnotesize{#1}}
\newcommand{\evdownnavi}{EDS}

\newcommand{\sampleRate}{prob_init}
\newcommand{\accAvg}{test_accuracy_avg}
\newcommand{\NCaltechAcc}{Test Accuracy-mean}
\newcommand{\genOneAPFifty}{test-AP_50-mean}
\newcommand{\psnr}{PSNR_avg-mean} 

\newcommand{\meanAe}{mean_ae-mean}
\newcommand{\lwidth}{0.25mm}
\newcommand{\scaleFactor}{100}
\newcommand{\plotSampler}[6]{ 
    \ifthenelse{\isempty{#6}}{ 
        \addplot table [
            col sep=comma, 
            x=prob_init, 
            y=#2, 
            restrict expr to domain={\thisrow{sampler_id}}{#3:#3},
        ] {#1};
    }{
        \addplot [#6] table [
            col sep=comma, 
            x=prob_init, 
            y=#2, 
            restrict expr to domain={\thisrow{sampler_id}}{#3:#3},
        ] {#1};
    }

    \ifthenelse{\isempty{#5}}{}{
        \addlegendentry{#5}
    }
}

\newcommand{\plotScaler}[3]{  
    \pgfplotstablecreatecol[
        create col/expr={\thisrow{#2} * #3} 
    ]{#2}{#1}
}

\definecolor{cvprblue}{rgb}{0.21,0.49,0.74}
\usepackage[pagebackref,breaklinks,colorlinks,allcolors=cvprblue]{hyperref}

\usepackage[capitalize]{cleveref}
\crefname{section}{Sec.}{Secs.}
\Crefname{section}{Section}{Sections}
\Crefname{table}{Table}{Tables}
\crefname{table}{Tab.}{Tabs.}

\def\paperID{2} 
\def\confName{CVPR}
\def\confYear{2025}

\title{Probabilistic Online Event Downsampling}

\author{Andreu Girbau-Xalabarder\\
Denso IT Laboratory\\
{\tt\small andreu@mail.d-itlab.co.jp}
\and
Jun Nagata\\
Denso IT Laboratory\\
\and
Shinichi Sumiyoshi\\
Denso IT Laboratory\\
\and
Ricard Marsal\\
National Institute of Informatics\\
\and
Shin'ichi Satoh\\
National Institute of Informatics
}

\begin{document}
\maketitle
\begin{abstract}

Event cameras capture scene changes asynchronously on a per-pixel basis, enabling extremely high temporal resolution. 
However, this advantage comes at the cost of high bandwidth, memory, and computational demands.
To address this, prior work has explored event downsampling, but most approaches rely on fixed heuristics or threshold-based strategies, limiting their adaptability. 
Instead, we propose a probabilistic framework, \alg, that models event importance through an event-importance probability density function (ePDF), which can be arbitrarily defined and adapted to different applications.
Our approach operates in a purely online setting, estimating event importance on-the-fly from raw event streams, enabling scene-specific adaptation.
Additionally, we introduce zero-shot event downsampling, where downsampled events must remain usable for models trained on the original event stream, without task-specific adaptation.
We design a contour-preserving ePDF that prioritizes structurally important events and evaluate our method across four datasets and tasks—object classification, image interpolation, surface normal estimation, and object detection—demonstrating that intelligent sampling is crucial for maintaining performance under event-budget constraints.
Code available \href{https://github.com/DensoITLab/POLED}{here}.

\end{abstract}

\begin{figure}[t]
    \centering
    \includegraphics[width=\linewidth]{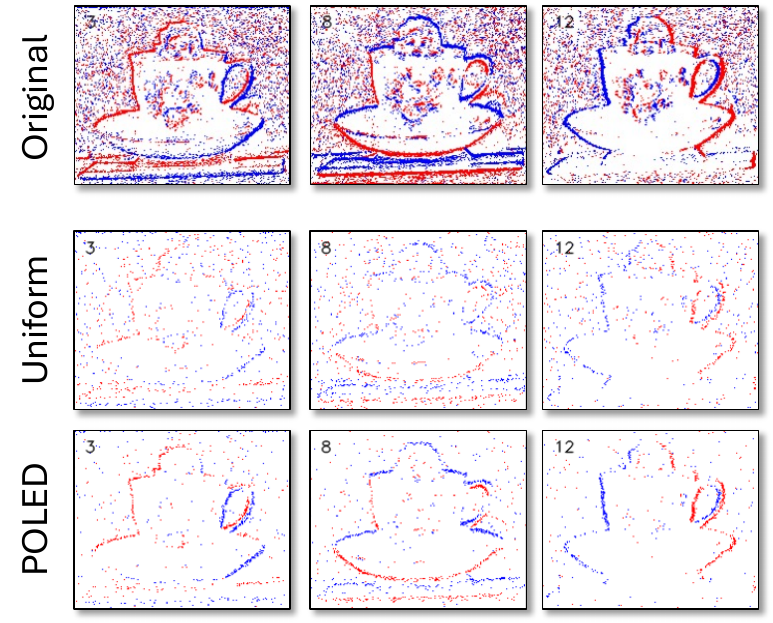}
    \caption{\textit{Cup} scene from N-Caltech101. By selectively retaining important events, \alg improves efficiency while maintaining task-relevant details, unlike naive uniform sampling.
    Top: Original event stream. Middle: Uniformly downsampled events. Bottom: Downsampled events with \alg, which adapts to the scene to preserve structured information, making contours more explicit.}
    \label{fig:teaser}
\end{figure}
    
\section{Introduction}

Event cameras, also known as neuromorphic cameras, differ fundamentally from conventional RGB sensors. 
Instead of capturing images at fixed frame rates, they asynchronously detect changes in intensity, producing a stream of events. 
Each event encodes its pixel location, precise timestamp (microsecond resolution), and polarity, indicating whether the intensity increased (\ON) or decreased (\OFF).
Key advantages of event cameras include a high dynamic range (up to 140 dB), low power consumption, no motion blur, high temporal resolution, and low latency. 
These properties make them well-suited for tasks such as video deblurring, high-speed object detection and tracking, and super-slow-motion video synthesis.

However, the sheer volume of events they generate poses significant challenges.
Processing, storing, and transmitting event streams efficiently remains a major bottleneck, particularly in embedded systems or bandwidth-constrained applications like ADAS (Advanced Driver Assistance Systems), or scenarios with limited data transmission capabilities.
To address this, event downsampling has emerged as a key strategy, aiming to reduce event rates while preserving the most relevant information in the scene.

To address this, most existing methods reduce the number of events by downscaling spatial dimensions or aggregating events temporally, typically using predefined thresholds. 
While effective in reducing data, these methods lack adaptability and do not consider the importance of individual events.

We take a different approach, formulating event downsampling as a stochastic process. 
Given an estimated event-importance probability density function (ePDF), incoming events are sampled probabilistically based on their likelihood of belonging to the distribution. 
This framework is adaptive, allowing the downsampling strategy to adjust dynamically to scene statistics or to the task at hand by modifying the ePDF on the fly.
To mimic realistic scenarios-where a camera is recording and bandwidth and computational resources are limited-, we adopt an online downsampling perspective, where only past and present events are available when making decisions. 
In contrast, offline methods can access future events, at the cost of increased memory usage and latency.

Additionally, most prior event downsampling methods have been evaluated on simplistic classification tasks, making it unclear how well they generalize to real-world applications. 
We argue that a truly effective downsampling strategy should be tested across multiple datasets and tasks, and evaluate our method on classification (NCaltech101~\cite{orchard2015converting}), surface normals estimation (ESFP~\cite{Muglikar23CVPR}), frame interpolation (Timelens~\cite{tulyakov2021time}), and object detection in automotive scenario for Gen1 Automotive Detection dataset~\cite{de2020large}.

Furthermore, we emphasize zero-shot event downsampling, where the downsampled event stream is directly fed into a pre-trained model without task-specific adaptation. 
This ensures broad applicability in scenarios where retraining is impractical or infeasible due to limited accessibility, computational resources, time, or data constraints.
To complement this, we also investigate the effects of retraining the models on downsampled events, providing insights into how adaptation impacts performance when retraining is an option.

In summary, our contributions are as follows:

\begin{itemize}
    \item We present a probabilistic online event downsampling framework (\alg), where events are sampled based on an arbitrarily defined event-importance probability density function (ePDF), allowing for adaptable and task-agnostic event reduction.
    \item We introduce an ePDF that prioritizes the preservation of edges, as well as two baselines.
    \item We evaluate our approach in a zero-shot setting, where downsampled events are directly fed into pre-trained models without task-specific adaptation, and validate our method on four diverse tasks—classification (N-Caltech101), object detection (Gen1 Automotive Detection), frame interpolation (Timelens), and surface normals estimation (ESFP)—demonstrating that intelligent event sampling is key to preserving performance in event-budget scenarios.
\end{itemize}

\section{Related work}

Modern event cameras produce an overwhelming number of events, posing significant challenges in bandwidth, computation, and memory. 
Efficient event stream management is therefore crucial, with downsampling emerging as a key strategy to reduce computational and bandwidth demands.

Early works explored event downsampling in spatial and temporal dimensions by scaling event coordinates and timestamps, adapting the sampling strategy to the dataset \cite{cohen2018spatial}. 
Later approaches refined this by integrating events over space and time using a counting strategy with refractory periods \cite{ghoshevdownsampling}.

Other methods take inspiration from biological neurons, reducing events based on the activation of multiple sensory unit layers \cite{barrios2018less}. 
Spiking Neural Networks (SNNs) have also been explored for downsampling, leveraging neuromorphic processing to optimize event retention \cite{gupta2020implementing,Gruel_2023_WACV,ghosh2023insect,rizzo2023neuromorphic}. 
Adaptive compression strategies, such as Huffman encoding that dynamically adjusts to bandwidth constraints, have also been proposed \cite{bisulco2020near}, along with pre-processing techniques that use non-uniform spatial sampling via 3D grids \cite{bi2019graph}.

Beyond computational efficiency, research has also examined how downsampling affects human perception of event streams. 
Studies have compared basic temporal and spatial filtering with more advanced SNN-based approaches to assess how sparsification impacts interpretability \cite{gruel2023frugal,Gruel_2023_WACV}.

Despite these advances, most existing methods rely on fixed heuristics or task-specific optimizations, limiting their adaptability across different applications.

In contrast to existing approaches, we formulate event downsampling as an online stochastic process, where events are sampled based on their likelihood of belonging to an estimated event distribution. 
This enables adaptive selection based on event importance and scene statistics, rather than relying on fixed heuristics or thresholds.

The closest work to ours is \cite{araghi2024pushing}, which uniformly samples events each epoch to train a CNN, studying the effects of event reduction and its interaction with CNN training parameters. 
However, their approach does not consider event importance or zero-shot applicability. 
Instead, we focus on the downsampling technique itself, evaluating its performance across independent tasks and models trained on the original event stream. 
Additionally, we investigate the effects of retraining with downsampled events, reaching similar conclusions to \cite{araghi2024pushing} while emphasizing the role of intelligent sampling.

We propose importance-based downsampling using an event-importance probability density function (ePDF), which can be arbitrarily defined and adapted to different settings. 
To make this framework broadly applicable, we introduce a generic formulation and processing pipeline, namely \alg, capable of handling any valid ePDF. 
In this work, we present a Poisson-based ePDF designed to prioritize contour preservation, under the premise that contour-related events are more relevant for solving diverse tasks.
Furthermore, we approach event downsampling from a purely online perspective, making decisions based only on past and present information, simulating a real-time scenario where future events are unknown.

Finally, prior work has largely focused on simple classification datasets or introduced metrics favoring classification-based evaluation. 
To provide a more comprehensive assessment, we evaluate our method on four challenging datasets, covering classification, frame interpolation for super-slow-motion video generation, surface normal estimation, and object detection in an automotive setting.

\begin{figure*}[t!]
    \centering
    \includegraphics[width=\linewidth]{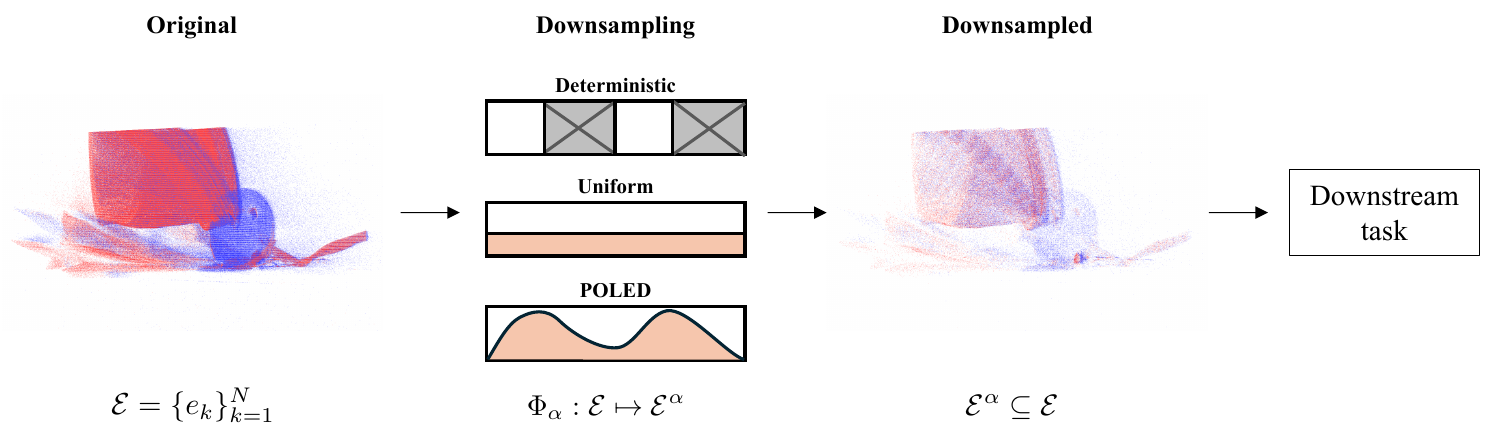}
    \caption{Overview of the proposed downsampling pipeline. 
    To simplify visualization, the downsampling process is illustrated for a single pixel over time rather than across the full spatial-temporal event grid.
    The original event stream $\evSO$ is processed using one of three introduced downsampling techniques: Deterministic, Uniform, or \alg (our probabilistic approach).
    Deterministic applies fixed temporal acceptance windows, Uniform assigns a constant acceptance probability to each event, and \alg estimates an event probability density function (ePDF) to sample events based on importance.
    The resulting downsampled event $\evSD$ stream is then used for downstream tasks, preserving the most relevant events while reducing data volume.}
    \label{fig:samplingMethods}
\end{figure*}

\section{Method}
\label{sec:method}

In this section, we formalize event downsampling as a stochastic process, present two baseline approaches, one deterministic and one based on uniform sampling, and introduce an adaptive Poisson-based event-importance probability density function (ePDF) designed to prioritize events near edges, which we hypothesize to be generally important across different tasks.

Overview of the proposed downsampling pipeline. The original event stream $\mathcal{E}$ is processed using one of three downsampling techniques: Deterministic, Uniform, or POLED (our probabilistic approach). The resulting downsampled event stream $\mathcal{E}_\omega$ is then fed into a downstream task model. The downsampling process is designed to retain the most relevant events while reducing data volume for computational efficiency.

\subsection{Problem formulation}
Event cameras operate asynchronously, responding independently to changes in light intensity at each pixel. 
A comparator monitors intensity variations relative to a reference level: if the increase surpasses a predefined threshold, an \ON\ event is generated; if the decrease exceeds the threshold, an \OFF\ event is triggered. 
No event is produced when intensity remains unchanged.

Each event is represented as:
\begin{equation}
    e_k=({\bf x}_k, t_k, p_k)~,~ k \in \mathbb{N}^+
\end{equation}
where ${\bf x}_k=[x_k, y_k]$ is the pixel location, $t_k$ denotes the timestamp at which the $k$-th event occurs, and $p_k \in \{\text{\ON{}}, \text{\OFF{}}\}$ specifies the polarity of the change.

A sequence of events, or an event stream, is defined as:
\begin{equation}
    \evSO=\{e_k\}^{N}_{k=1}
\end{equation}

We define event downsampling as the problem of finding a mapping $\fDown$ such that:
\begin{equation}
    \fDown : \evSO \mapsto \evSD, \quad \evSD \subseteq \evSO   
\end{equation}
\begin{equation}
    \begin{aligned}
        \fDown(e_k) &= 
        \begin{cases}
        e_k, & \quad \text{if condition holds} \\
        \emptyset, & \quad \text{otherwise}
        \end{cases}
    \end{aligned}
\end{equation}
where $\evSD$ is the downsampled event stream, and $\alpha \in (0, 1)$ denotes the downsampling ratio with respect to the original event set $\evSO$.
Note that we define $\evSD$ as a subset of the original events $\evSO$, meaning that no new events are generated -nor are existing ones modified— only a selection of events from the original set is removed. 
This is in contrast to previous works, where event information may be altered, aggregated, or even synthesized.

\subsection{Online Event Downsampling}

To determine whether an incoming event \ek\ is accepted or rejected, we explicitly adopt an \textit{online} downsampling approach, meaning that only past and present events in \evSOTnT{1}{k} are available at any given time. 
In contrast, an \textit{offline} setting would allow access to future events, i.e., decisions regarding \ek\ could be influenced by information in \evSOTnT{1}{l}, where $l > k$.  
A practical example of offline downsampling occurs in methods that aggregate events within a temporal window extending beyond the current event \ek\ .

\minisection{Enforcing event rate constraints.}
To ensure that downsampling adheres to a predefined event budget—modeled as the downsampling ratio $\alpha$—we introduce an event capping mechanism.
This mechanism enforces the constraint that the number of retained events must not exceed the allowed proportion $\alpha$ of the original stream, simulating real-world constrains on bandwidth or computational resources.

Formally, given the incoming event \ek\, we define the downsampled event relation as:
\begin{equation}
    r = \frac{|\evSDTnTneq{0}{k-1}|}{|\evSOTnTneq{0}{k-1}|}
\end{equation}
where $|\cdot|$ is the dimension of the set. 
If at any point $r > \alpha$, the event \ek\ is dropped, even if it would have otherwise been accepted by the sampling method.

\subsection{Baselines}
\label{sec:baselines}

To assess the effectiveness of our probabilistic event downsampling approach, we introduce two baseline strategies: deterministic and uniform downsampling. 

\subsubsection{Deterministic Downsampling}
The first approach we explore is a deterministic downsampling strategy, where small bursts of events are retained based on their timestamps within predefined temporal acceptance windows.

To achieve this, we define a small sliding temporal window $T_w$ and partition it into two sub-intervals: an acceptance window $T_a$ and a rejection window $T_r$, where $T_a = T_w \cdot \alpha$ and $T_r = T_w - T_a$.
In practice, we use $T_w=0.1$ milliseconds.
The downsampling process is anchored at an initial reference time $T_{0}$, which shifts as $T_{0}=nT_w$, defining the start of the $n$-th temporal window.
An event $e_k = ({\bf x}_k, t_k, p_k)$ is retained if its timestamp falls within the acceptance window, defined as :
\begin{equation}
    \fDown(e_k) =
    \begin{cases}
        e_k, & \quad \text{if} \quad T_0 < t_k \leq T_0 + T_a  \\
        \emptyset, & \quad \text{if} \quad T_0 + T_a < t_k \leq T_0 + T_w
    \end{cases}  
\end{equation}

\subsubsection{Uniform Downsampling}
The second baseline follows a uniform random sampling strategy, where events are accepted independently with a fixed probability $\alpha$. 
Formally:
\begin{equation}
    \begin{aligned}
        \fDown(e_k) &= 
        \begin{cases}
        e_k, & \quad \text{with probability } \alpha \\
        \emptyset, & \quad \text{with probability } 1 - \alpha
        \end{cases}
    \end{aligned}
\end{equation}

This baseline represents the simplest stochastic approach to event downsampling and serves as a reference point for evaluating our more structured probabilistic methods.

\subsection{Probabilistic Online Event Downsampling}

We now introduce Probabilistic Online Event Downsampling (\alg), a flexible framework for event reduction based on probabilistic sampling. 
Instead of applying fixed rules or uniform selection, this approach models event importance using an estimated event Probability Density Function (ePDF). 
Given any valid ePDF \pdf{}, each incoming event \ek\ is accepted or rejected based on its likelihood of belonging to the estimated event distribution.

To achieve this, we define the mapping $\fDown$ as:
\begin{equation}
    \begin{aligned}
        \fDown(e_k) &= 
        \begin{cases}
        e_k, & \quad \text{with probability } \pa \\
        \emptyset, & \quad \text{with probability } 1 - \pa
        \end{cases}
    \end{aligned}
\end{equation}
where $\pa$ represents the probability of accepting an event $e_k = ({\bf x}_k, t_k, p_k)$ derived from the ePDF.

\minisection{Pipeline.}
The downsampling pipeline is as follows. 
First, we estimate the event probability density function (ePDF), denoted as \pdfSOTnT{1}{k}.
The way to estimate this PDF depends on how the event importance is defined.
In this work, we prioritize events clustered around edges (Sec.~\ref{sec:poisson}), as they are generally more informative across different contexts.

To focus on recent changes in the scene, we estimate the ePDF within a sliding temporal window of size $T$, using events from the interval \pdfSOTnT{(n-1)T}{nT}.
Following this estimation, we evaluate the likelihood of an incoming event \ek\, for $t_k > nT$, belonging to the estimated distribution \pdfSO{}.
For clarity, we drop the explicit temporal notation and refer to the estimated event distribution as \pdfSO{}.

To convert the likelihood of an event \ek\ into an acceptance probability, we define a mapping function $\score$ that transforms the estimated likelihood \pdfSO{} into a probability score:
\begin{equation}
    \pa = \score(\pdfSOneq(\mathbf{x}_k))
    \label{eq:score}
\end{equation}

The resulting probability $\pa$ is then used as the acceptance probability in a Bernoulli sampling process.

To ensure that $\pa$ is in the valid range $[0, 1]$, the mapping function $\score$ is defined as:
\begin{equation}
    \score(\mathbf{x}) = \sigma\left( \pdfSOneq^{*}(\mathbf{x}) + (\alpha - \overline{\pdfSOneq^{*}}(\mathbf{x}))~;~\theta \right)
    \label{eq:scaling}
\end{equation}
where $\pdfSOneq^{*}$ is the normalized form of $\pdfSOneq$ by the min-max normalization, its mean value displaced to match the downsampling ratio $\alpha$, and its values rescaled by the sigmoid function $\sigma$ to scale it to a probability range $[0,1]$.

We define the $\theta$ parameters to scale and shift the sigmoid function, ensuring that no event has a probability of exactly $0$ (allowing all events a chance of being selected) or exactly $1$ (allowing even the most likely events to be removed occasionally).
\begin{equation}
    \sigma(x; \theta) = \frac{1}{1 + e^{-\theta_{1} (x - \theta_{2})}}
\end{equation}
In this work, we set $\theta = [\theta_1, \theta_2] = [5, 0.5]$, ensuring that mid-range values follow a near-linear probability scaling, while extreme values are adjusted to prevent hard acceptance or rejection.

\minisection{On the importance of events.}
Defining a generic notion of event importance is inherently ill-posed, as relevance depends on the specific objective at hand.
Ideally, our goal is to define a \pdfSO{} that is task-agnostic and generalizable across diverse applications.
To this end, in the next section, we introduce an approach that prioritizes edge events, as they tend to be informative across a wide range of vision tasks.
However, edge-based downsampling alone may overlook task-specific event distributions.
For instance, in object detection, events corresponding to objects of interest—such as cars or pedestrians—are more valuable than background events.

To bridge this gap, we propose incorporating task-specific prior knowledge into the event downsampling process.
This is achieved by integrating a prior distribution that modulates the estimated ePDF, introducing a bias that ensures the selection process aligns with task-relevant event patterns.
Whenever a task-specific prior is available, we modify \cref{eq:scaling} (omitting $\mathbf{x}$ for clarity) as:
\begin{equation}
    \score(\mathbf{x}) = \sigma\left( (\pdfSOneq \cdot \pdfP)^{*} + (\alpha - \overline{(\pdfSOneq \cdot \pdfP)^{*}})~;~\theta \right)
    \label{eq:scalingprior}
\end{equation}
where $\pdfP$ is the task-specific prior distribution, and $(\pdfSOneq \cdot \pdfP)^{*}$ is the min-max normalized form of the product of the two distributions.
The prior $\pdfP$ can be derived from dataset statistics, heuristically defined, or learned from data, depending on the task requirements.

\subsubsection{Poisson-based Event PDF}
\label{sec:poisson}
While event significance varies across applications, we argue that events that correspond to edges are generally informative across diverse tasks.
Building on this intuition, we introduce a Poisson-based method for estimating the ePDF, designed to prioritize edge-preserving downsampling.

Poisson distributions are widely used to model the occurrence of discrete events in space and time, making them well-suited for capturing the sparsity and stochastic nature of event data.
Given a rate parameter $\lambda$ representing the expected number of events in a given region, the probability mass function (PMF) of a Poisson-distributed variable $k$ is:
\begin{equation}
    P(k; \lambda) = \frac{\lambda^k e^{-\lambda}}{k!}
\end{equation}

In our case, we estimate an event probability density function (ePDF) using a spatial Poisson process, where $\lambda$ is dynamically computed based on pixel-wise event densities within a temporal window $T$.
This allows us to prioritize edge-preserving downsampling by assigning higher probabilities to regions with structured event activity.

To model event importance, we define the ePDF as the probability of at least one event occurring at a given spatial location $\mathbf{x}$ within the temporal window $[(n-1)T, nT]$:
\begin{equation}
    \pdfSOneq(\mathbf{x}) = 1 - P(0; \lambda(\mathbf{x})) = 1 - e^{-\lambda(\mathbf{x})}
\end{equation}
where $\lambda(\mathbf{x})$ represents the event density at $\mathbf{x}$ over the temporal window $T$.

Finally, we convert \pdfSO{} into an acceptance probability using Eq. \ref{eq:score}, ensuring that the sampling process prioritizes edge events while dynamically adapting to scene variations.

\begin{table*}[t]
    \centering
    \setlength{\tabcolsep}{6mm}
    \resizebox{\linewidth}{!}{
    \begin{tabular}{@{}lccccccc@{}}
    \toprule
        & Events & N-Caltech101~\cite{orchard2015converting} & Gen1~\cite{de2020large}  & \multicolumn{2}{c}{Timelens~\cite{tulyakov2021time}} & ESFP-Real~\cite{Muglikar23CVPR} \\
    \cmidrule(lr){3-3}
    \cmidrule(lr){4-4}  
    \cmidrule(lr){5-6}
    \cmidrule(lr){7-7}
            &  & Accuracy~$\uparrow$ &	AP50~$\uparrow$   &	\textit{Close} & \textit{Far} & AE~$\downarrow$  \\
    \midrule
    Original	    & 100\%	&	84.4	& 68.8	&	$32.47 \pm 4.6$	&	$32.04 \pm 1.8$	&	28.50	\\
    \midrule
    \evdownnavi~\cite{ghoshevdownsampling}	& 5\%	& 5.2 &	0.4	&	$29.58 \pm 5.0$	&	$24.10 \pm 1.6$	&	30.05	\\																							
    Deterministic	& 5\%	& 13.4 &	7.4	&	$29.75 \pm 5.0$	&	$24.48 \pm 1.7$	&	29.92	\\
    Uniform	        & 5\%	& 12.9 &	8.2	&	$30.23 \pm 4.6$	&	$26.03 \pm 2.1$	&	29.90	\\
    \algpoi	        & 5\%	& \textbf{27.4} &	\textbf{10.3}	&	\textbf{$30.39 \pm 4.6$}	&	\textbf{$26.87 \pm 2.4$}	&	\textbf{29.84}	\\
    \midrule
    \evdownnavi	& 10\%	& 23.7 &	2.6	&	$29.75 \pm 5.0$	&	$24.47 \pm 1.6$	&	29.76	\\
    Deterministic	& 10\%	& 41.5 &	18.7	&	$29.96 \pm 5.0$	&	$24.66 \pm 1.7$	&	29.51	\\
    Uniform	        & 10\%	& 40.3 &	19.9	&	$30.89 \pm 4.5$	&	$27.45 \pm 2.0$	&	29.52	\\
    \algpoi	        & 10\%	& \textbf{55.2} &	\textbf{24.3}	&	\textbf{$30.99 \pm 4.5$}	&	\textbf{$28.23 \pm 2.3$}	&	\textbf{29.47}	\\
    \midrule																																										
    \evdownnavi	& 50\%	& 77.9 &	61.6	&	$30.97 \pm 4.6$	&	$27.37 \pm 2.3$	&	28.57	\\
    Deterministic	& 50\%	& 82.7 &	60.5	&	$31.31 \pm 4.6$	&	$27.90 \pm 2.2$	&	28.46	\\
    Uniform	        & 50\%	& 82.9 &	61.1	&	$32.05 \pm 4.5$	&	$30.80 \pm 2.0$	&	\textbf{28.45}	\\
    \algpoi	        & 50\%	& \textbf{83.3} &	\textbf{61.8}	&	\textbf{$32.07 \pm 4.5$}   &	\textbf{$30.82 \pm 2.1$}   &	28.46	\\
    \bottomrule
    \end{tabular}
    }
    \vspace*{-1mm}
    \caption{Results on a zero-shot setting of the four approaches for different levels of downsampling. For Uniform and \algpoi average of several runs. Best in \textbf{bold}.}
    \label{tab:res_all}
\end{table*}

\section{Experiments}
\label{sec:experiments}

We evaluate our proposed probabilistic event downsampling method (\alg) across four diverse datasets, each representing a distinct vision task.
Our goal is to assess both the generalization capabilities of our approach—particularly the Poisson-based ePDF defined in \cref{sec:poisson}, named as \algpoi in the results—and its impact on downstream performance across various applications.

Our experiments cover the following datasets and tasks:
\begin{itemize}
    \item \textbf{N-Caltech101}~\cite{orchard2015converting} for object classification, using the model from~\cite{Gehrig_2019_ICCV}.
    \item \textbf{Gen1}~\cite{de2020large} for object detection in driving scenarios, leveraging the approach from~\cite{gehrig2023recurrent}.
    \item \textbf{Timelens}~\cite{tulyakov2021time} for frame interpolation, evaluated with its original model.
    \item \textbf{ESFP}~\cite{Muglikar23CVPR} for surface normal estimation, with the method introduced in the same work.
\end{itemize}

These datasets present a diverse set of challenges, allowing us to evaluate the robustness of our downsampling strategy beyond standard classification benchmarks.
We compare our method against our deterministic and uniform downsampling baselines, introduced in \cref{sec:baselines}, as well as a recently proposed event downsampling approach EDS~\cite{ghoshevdownsampling}, analyzing performance in both zero-shot scenarios and retraining settings when applicable.

\minisection{\evdownnavi\ \cite{ghoshevdownsampling}.}
To provide a broader comparison, we include \evdownnavi (Event DownSampling), a recent method that simulates a lower-resolution event camera by reducing both spatial and temporal dimensions.

While originally designed for a different purpose, \evdownnavi\ offers an interesting point of comparison: it represents an alternative downsampling strategy—one that naively reduces resolution instead of selecting events based on importance.
By evaluating its performance in event-budget scenarios, we can empirically assess whether simple spatio-temporal downsampling is sufficient, or if intelligent event selection is necessary to retain task-critical information.

\subsection{Implementation details}
We evaluate our approach using existing methods tailored to each dataset and task.
For N-Caltech101 and ESFP-Real, we train the classifiers from \cite{Gehrig_2019_ICCV} and \cite{Muglikar23CVPR} respectively on the original event streams, as well as on the downsampled event streams (Sec.~\ref{sec:results_retraining}).
For Timelens and Gen1, we use the pretrained models and weights provided in~\cite{tulyakov2021time} and~\cite{gehrig2023recurrent}, respectively. 
Both models are trained on the original event streams and zero-shot tested on the downsampled event streams.

To enforce an online processing constraint, we sequentially load datasets and process events one-by-one.
All downsampling methods operate with a temporal window of $6~\text{ms}$, approximately corresponding to 160 FPS.
Downsampling is performed entirely on the CPU, while all method-related experiments were conducted on a single NVIDIA Quadro RTX 8000 GPU.

\subsection{Introductory results}
\label{sec:results_into}
Table~\ref{tab:res_all} presents the zero-shot performance of the different downsampling techniques accross all datasets and tasks.
We report the original performance using 100\% of the event stream, along with the zero-shot performance at 5\%, 10\%, and 50\% from the original events.

At low-event budget scenarios ($5\%$ and $10\%$), we observe that \algpoi, the Poisson-based ePDF method, outperforms the deterministic and uniform downsampling baselines in all datasets and tasks, by a considerable margin in some cases such as in N-Caltech101 and Gen1.
While events naturally cluster around edges, our results suggest that further emphasizing edge information is crucial for maintaining performance, particularly in low-event regimes.
At higher event budgets (50\%), performance differences among methods diminish, as essential task-relevant events are naturally retained due to their inherent concentration around contours.

Among all techinques, \evdownnavi~\cite{ghoshevdownsampling} consistently performs the worst across all datasets and tasks.
We attribute this to three key factors.
First, the method applies spatial downsampling, reducing resolution before passing the events to the downstream model. 
Since our evaluation setup assumes a method-agnostic downsampling process, events must be rescaled back to the original resolution, potentially introducing artifacts and loss of fine-grained details.
Second, \evdownnavi\ integrates events within fixed temporal intervals, which may not capture the temporal dynamics of the events accurately, particularly in low-event regimes.
Finally, the combined effects of spatial and temporal aggregation may alter the statistical distribution of the event stream, introducing a distribution shift that impacts downstream performance.

\begin{figure}[t]
    \centering
    \begin{subfigure}{0.48\linewidth}
        \centering
        \includegraphics[width=\linewidth]{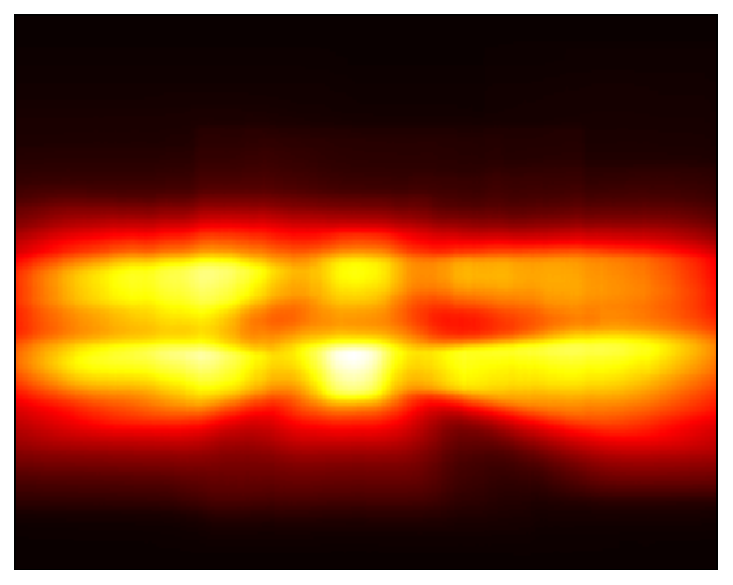}
        \caption{Objects distribution}
        \label{fig:RVT_priors:prior}
    \end{subfigure}
    \hfill
    \begin{subfigure}{0.48\linewidth}
        \centering
        \includegraphics[width=\linewidth]{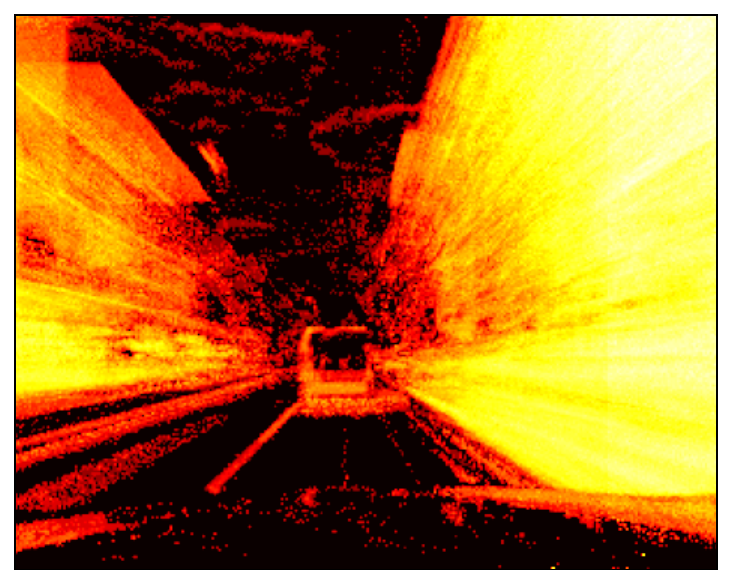}
        \caption{Events distribution}
        \label{fig:RVT_priors:orig}
    \end{subfigure}
    \caption{Comparison of the distribution of a) bounding boxes in the training set and b) raw event distribution in the Gen1 dataset.}
    \label{fig:RVT_Priors}
\end{figure}

\subsection{Event importance}
\label{sec:results_importance}
As previously discussed, not all events contribute equally to solving a task. While event importance is inherently task-dependent, our results suggest that events concentrated around object contours generally play a crucial role across a wide range of tasks.
However, not all event concentrations are equally important. 
While in Timelens most contours contribute equally to solve the task, in the Gen1 dataset for car and pedestrian detection, events concentrated around vehicle edges are far more critical than those outlining sidewalks.

We illustrate this in Figure~\ref{fig:RVT_Priors}.
Comparing the distribution of training set bounding boxes (Fig.\ref{fig:RVT_priors:prior}) with the raw event distribution (Fig.\ref{fig:RVT_priors:orig}), we observe a significant mismatch between event locations and actual detections.
This suggests that \algpoi, which prioritizes high-density regions, may oversample events in areas like trees on sidewalks rather than focusing on the most relevant contours.

To address this misalignment, we incorporate the training set bounding box distribution as a prior, using \cref{eq:scalingprior} to modulate the ePDF and guide sampling toward the task-relevant regions.
Also, in N-Caltech101, we use a 2D Gaussian distribution as the objects of interest tend to concentrate at the center of the image.
Table~\ref{tab:res_prior} presents the performance of \algpoi with and without a task-specific prior, highlighting the performance gains achieved by incorporating prior knowledge.

\begin{table}[t]
    \centering
    \setlength{\tabcolsep}{1.5mm}
    \resizebox{\linewidth}{!}{
    \begin{tabular}{@{}lccccc@{}}
    \toprule
        & Events & \multicolumn{2}{c}{N-Caltech101 (Accuracy~$\uparrow$)} & \multicolumn{2}{c}{Gen1 (AP50~$\uparrow$)}  \\
    \cmidrule(lr){3-4}
    \cmidrule(lr){5-6}
                    &  & w/o prior & w/ prior & w/o prior & w/ prior \\
    \midrule
    Original	    & 100\%	&	84.4	& - & 68.8 & - \\
    \midrule
    \algpoi	        & 5\%	&	25.1 & \best{27.4} (+2.3) &	9.3 & \best{10.3} (+1.0) \\
    \midrule
    \algpoi	        & 10\%	&	52.0 & \best{55.2} (+3.2) &	22.1 & \best{24.3} (+2.2) \\
    \midrule																																										
    \algpoi	        & 50\%	&	83.1 & \best{83.3} (+0.2) &	60.9 & \best{61.8} (+0.9) \\
    \bottomrule
    \end{tabular}
    }
    \vspace*{-1mm}
    \caption{Zero-shot performance of \alg (Poisson-based) downsampling on N-Caltech101 and Gen1, comparing results with and without a task-specific prior.}
    \label{tab:res_prior}
\end{table}

\subsection{Task specific re-training}
\label{sec:results_retraining}

\pgfplotstableread[col sep=comma]{data/NCaltech101_data.csv}\datatable
\pgfplotstableread[col sep=comma]{data/NCaltech101_retrained_data.csv}\datatableRetrained

\plotScaler{\datatable}{\NCaltechAcc}{\scaleFactor}
\plotScaler{\datatable}{\sampleRate}{\scaleFactor}
\plotScaler{\datatableRetrained}{\NCaltechAcc}{\scaleFactor}
\plotScaler{\datatableRetrained}{\sampleRate}{\scaleFactor}

\pgfmathsetmacro{\xmin}{0}
\pgfmathsetmacro{\xmax}{1}
\pgfmathsetmacro{\ymin}{0}
\pgfmathsetmacro{\ymax}{0.9}
\pgfmathsetmacro{\domainmin}{0.001}
\pgfmathsetmacro{\domainmax}{1}

\pgfmathsetmacro{\xminScaled}{\xmin * \scaleFactor}
\pgfmathsetmacro{\xmaxScaled}{\xmax * \scaleFactor}
\pgfmathsetmacro{\yminScaled}{\ymin * \scaleFactor}
\pgfmathsetmacro{\ymaxScaled}{\ymax * \scaleFactor}
\pgfmathsetmacro{\domainminScaled}{\domainmin * \scaleFactor} 
\pgfmathsetmacro{\domainmaxScaled}{\domainmax * \scaleFactor} 

\pgfmathsetmacro{\baseline}{0.844}
\pgfmathsetmacro{\baselineScaled}{\baseline * \scaleFactor}

\begin{figure}[t]
    \centering
    \begin{tikzpicture}
        \begin{axis}[
            width=\columnwidth,
            xlabel={Sampling rate (\%)},
            ylabel={Accuracy},
            xmin=\xminScaled,
            xmax=\xmaxScaled,
            ymin=\yminScaled,
            ymax=\ymaxScaled,
            xmode=log,
            log basis x=10,
            log ticks with fixed point,
            xtick pos=left, 
            ytick pos=left,  
            domain=\domainminScaled:\domainmaxScaled, 
            legend style={
                at={(0.5,1.05)}, 
                anchor=south, 
                legend columns=-1, 
                /tikz/every even column/.append style={column sep=5pt} 
            },
            legend image post style={
                scale=1.5, 
                only marks
            }, 
        ]

            \plotSampler{\datatableRetrained}{\NCaltechAcc}{4}{\evdownnavi}{\evdownnavi}{smooth, mark=triangle, color=blue, line width=\lwidth}
            \plotSampler{\datatableRetrained}{\NCaltechAcc}{1}{Deterministic}{Det}{smooth, mark=diamond, color=orange, line width=\lwidth}
            \plotSampler{\datatableRetrained}{\NCaltechAcc}{2}{Uniform}{Uni}{smooth, mark=star, color=cyan, line width=\lwidth}
            \plotSampler{\datatableRetrained}{\NCaltechAcc}{3}{Poisson}{\algpoi}{smooth, mark=square, color=red, line width=1.5*\lwidth}



            \plotSampler{\datatable}{\NCaltechAcc}{4}{\evdownnavi}{}{mark=triangle, color=blue, line width=\lwidth, only marks}
            \plotSampler{\datatable}{\NCaltechAcc}{1}{Deterministic}{}{mark=diamond, color=orange, line width=\lwidth, only marks}
            \plotSampler{\datatable}{\NCaltechAcc}{2}{Uniform}{}{mark=star, color=cyan, line width=\lwidth, only marks}
            \plotSampler{\datatable}{\NCaltechAcc}{3}{Poisson}{}{mark=square, color=red, line width=\lwidth, only marks}
            
            \plotSampler{\datatable}{\NCaltechAcc}{4}{\evdownnavi}{}{smooth, mark=triangle, color=blue, line width=\lwidth, dashed}
            \plotSampler{\datatable}{\NCaltechAcc}{1}{Deterministic}{}{smooth, mark=diamond, color=orange, line width=\lwidth, dashed}
            \plotSampler{\datatable}{\NCaltechAcc}{2}{Uniform}{}{smooth, mark=star, color=cyan, line width=\lwidth, dashed}
            \plotSampler{\datatable}{\NCaltechAcc}{3}{Poisson}{}{smooth, mark=square, color=red, line width=1.5*\lwidth, dashed}

            \addplot[mark=none, dotted, black, samples=2, forget plot, line width=2*\lwidth] {\baselineScaled};

            \node at (axis cs:\domainminScaled,\baselineScaled) {\textcolor{black}{\Huge $\star$}};
            \node at (axis cs:\domainminScaled,\baselineScaled) [anchor=north west] {\textbf{Objective}}; 


        \end{axis}
    \end{tikzpicture}
    \captionsetup{skip=2pt} 
    \caption{Accuracy~($\uparrow$) vs. sampling rate on N-Caltech101~\cite{orchard2015converting} for zero-shot (dashed) and re-trained (solid) models at different downsampling levels. The purple star indicates the original performance. The x-axis is in logarithmic scale.}
    \label{fig:caltech}
\end{figure}
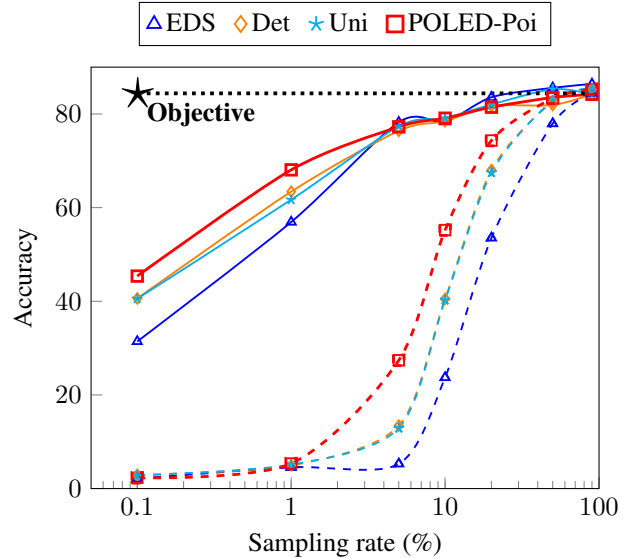
\pgfplotstableread[col sep=comma]{data/esfp_data.csv}\datatable
\pgfplotstableread[col sep=comma]{data/esfp_retrained_data.csv}\datatableRetrained

\plotScaler{\datatable}{\meanAe}{1}
\plotScaler{\datatable}{\sampleRate}{\scaleFactor}
\plotScaler{\datatableRetrained}{\meanAe}{1}
\plotScaler{\datatableRetrained}{\sampleRate}{\scaleFactor}

\pgfmathsetmacro{\xmin}{0}
\pgfmathsetmacro{\xmax}{1}
\pgfmathsetmacro{\ymin}{0.28}
\pgfmathsetmacro{\ymax}{0.307}
\pgfmathsetmacro{\domainmin}{0.001}
\pgfmathsetmacro{\domainmax}{1}

\pgfmathsetmacro{\xminScaled}{\xmin * \scaleFactor}
\pgfmathsetmacro{\xmaxScaled}{\xmax * \scaleFactor}
\pgfmathsetmacro{\yminScaled}{\ymin * \scaleFactor}
\pgfmathsetmacro{\ymaxScaled}{\ymax * \scaleFactor}
\pgfmathsetmacro{\domainminScaled}{\domainmin * \scaleFactor} 
\pgfmathsetmacro{\domainmaxScaled}{\domainmax * \scaleFactor} 

\pgfmathsetmacro{\baseline}{28.5}
\pgfmathsetmacro{\baselineScaled}{\baseline * 1}

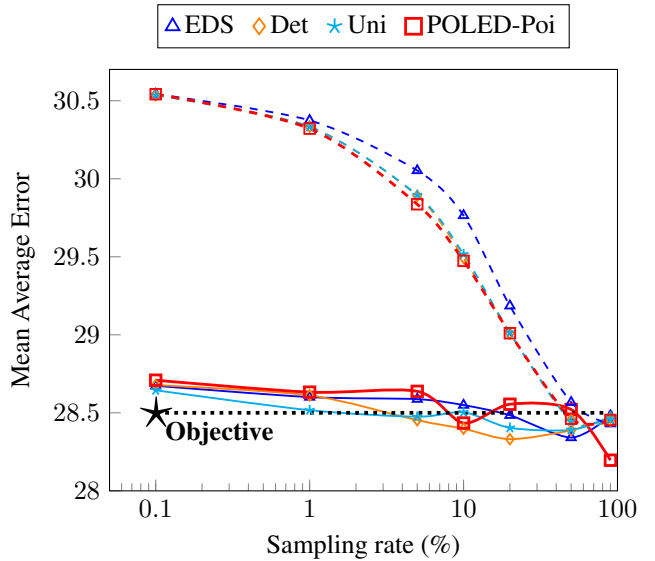
\begin{figure}[h!]
    \centering
    \begin{tikzpicture}
        \begin{axis}[
            width=\columnwidth,
            xlabel={Sampling rate (\%)},
            ylabel={Mean Average Error},
            xmin=\xminScaled,
            xmax=\xmaxScaled,
            ymin=\yminScaled,
            ymax=\ymaxScaled,
            xmode=log,
            log basis x=10,
            log ticks with fixed point,
            xtick pos=left, 
            ytick pos=left,  
            domain=\domainminScaled:\domainmaxScaled, 
            legend style={
                at={(0.5,1.05)}, 
                anchor=south, 
                legend columns=-1, 
                /tikz/every even column/.append style={column sep=5pt} 
            },
            legend image post style={
                scale=1.5, 
                only marks
            }, 
        ]

            \plotSampler{\datatableRetrained}{\meanAe}{4}{\evdownnavi}{\evdownnavi}{smooth, mark=triangle, color=blue, line width=\lwidth}
            \plotSampler{\datatableRetrained}{\meanAe}{1}{Deterministic}{Det}{smooth, mark=diamond, color=orange, line width=\lwidth}
            \plotSampler{\datatableRetrained}{\meanAe}{2}{Uniform}{Uni}{smooth, mark=star, color=cyan, line width=\lwidth}
            \plotSampler{\datatableRetrained}{\meanAe}{3}{Poisson}{\algpoi}{smooth, mark=square, color=red, line width=1.5*\lwidth}
            
            \plotSampler{\datatable}{\meanAe}{4}{\evdownnavi (Retrained)}{}{mark=triangle, color=blue, line width=\lwidth, only marks}
            \plotSampler{\datatable}{\meanAe}{1}{Deterministic (Retrained)}{}{mark=diamond, color=orange, line width=\lwidth, only marks}
            \plotSampler{\datatable}{\meanAe}{2}{Uniform (Retrained)}{}{mark=star, color=cyan, line width=\lwidth, only marks}
            \plotSampler{\datatable}{\meanAe}{3}{Poisson (Retrained)}{}{mark=square, color=red, line width=\lwidth, only marks}
            
            \plotSampler{\datatable}{\meanAe}{4}{\evdownnavi (Retrained)}{}{smooth, color=blue, line width=\lwidth, dashed}
            \plotSampler{\datatable}{\meanAe}{1}{Deterministic (Retrained)}{}{smooth, color=orange, line width=\lwidth, dashed}
            \plotSampler{\datatable}{\meanAe}{2}{Uniform (Retrained)}{}{smooth, color=cyan, line width=\lwidth, dashed}
            \plotSampler{\datatable}{\meanAe}{3}{Poisson (Retrained)}{}{smooth, color=red, line width=1.5*\lwidth, dashed}

            \addplot[mark=none, dotted, black, samples=2, forget plot, line width=2*\lwidth] {\baselineScaled};

            \node at (axis cs:\domainminScaled,\baselineScaled) {\textcolor{black}{\Huge $\star$}};
            \node at (axis cs:\domainminScaled,\baselineScaled) [anchor=north west] {\textbf{Objective}}; 


        \end{axis}
    \end{tikzpicture}
    \captionsetup{skip=2pt}
    \caption{Mean Average Error~($\downarrow$) vs. sampling rate on ESFP-Real~\cite{Muglikar23CVPR} for zero-shot (dashed) and re-trained (solid) models at different downsampling levels. The purple star indicates the original performance. The x-axis is in logarithmic scale.}
    \label{fig:esfp}
\end{figure}

While zero-shot event downsampling is ideal for real-world deployment—where re-training may not always be feasible—the performance gap between downsampled and full-event models remains significant.
To better understand the upper bound of downsampling effectiveness, we re-train models on the downsampled event streams for N-Caltech101 and ESFP, allowing them to fully adapt to the reduced event distribution.

Figures \ref{fig:caltech} and \ref{fig:esfp} compare zero-shot (dashed) vs. re-trained (solid) performance across different downsampling rates.
Notably, in both cases, the amount of events required to solve the task (i.e., achieving full-event performance) can be reduced dramatically:
on N-Caltech101, accuracy recovers almost fully with as little as 5\% of the original events, and on ESFP, nearly identical performance is achieved with just 0.1\% of the original event stream.
This aligns with findings from \cite{araghi2024pushing}, which also demonstrate that models can adapt to highly reduced event streams with appropriate training.

It is also important to note that the assumption that edge-related events are crucial holds in N-Caltech101 (Fig.~\ref{fig:caltech}), where the Poisson-based ePDF outperforms the baselines when both training and testing on downsampled events.
Additionally, at higher event retention rates (90\%), we observe that downsampled models exceed the original full-event performance, suggesting that our approach introduces a denoising effect, even though it was not explicitly designed for that purpose.

\subsection{Run time}
Computational efficiency is critical for event-based processing, particularly in real-time applications. 
Table~\ref{tab:res_runtime} reports the runtime of our proposed \algpoi\ compared to the deterministic and uniform baselines.
As expected, estimating the ePDF introduces additional overhead, making \algpoi\ approximately 1.4 times slower than the baselines. 
However, our current implementation is written in pure Python with a single blocking thread, representing a worst-case scenario for all techniques. 
Optimizing the implementation in lower-level languages such as C++ or CUDA would significantly improve runtime across all downsampling methods.
Additionally, \alg can benefit from parallelization, as the ePDF computation can be executed in a separate thread, reducing the impact of its overhead.

\begin{table}[t]
    \centering
    \setlength{\tabcolsep}{2mm}
    \resizebox{\linewidth}{!}{
    \begin{tabular}{@{}lccc@{}}
    \toprule
    Method          & Total (ms/Kev)    & PDF (ms/Kev)      & Eval (ms/Kev) \\
    \midrule
    Deterministic   & 4.90        & -                      & 0.45  \\
    Uniform         & 4.56        & -                      & 0.14  \\
    \algpoi         & 6.55        & 1.72                   & 0.14  \\
    \bottomrule
    \end{tabular}
    }
    \caption{Computational time per thousand events (ms/Kev) for different downsampling approaches. 
    \textit{Total} represents the overall downsampling time, \textit{PDF} accounts for computing the event probability density function, and \textit{Eval} measures the cost of the acceptance or rejection decision.}
    \label{tab:res_runtime}
\end{table}

\subsection{Qualitative results}
\begin{figure}[t]
    \centering
    \includegraphics[width=0.85\linewidth]{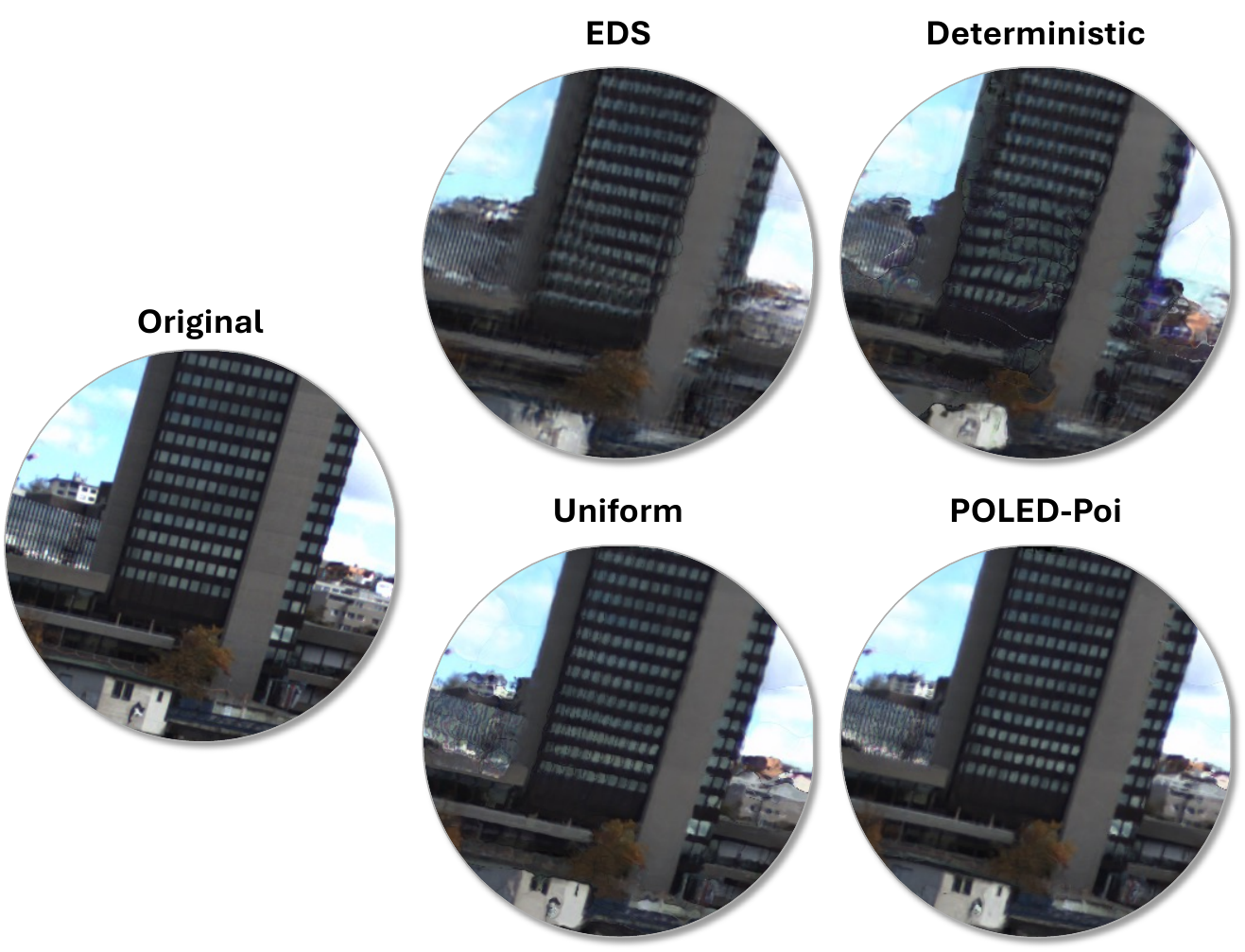}
    \caption{Detailed output of the frame interpolation method \cite{Gehrig_2019_ICCV} by using the event stream downsampled to maintain $10\%$ of the original events. Even at low-event regimes, \algpoi is able to keep the contour information (windows).}
    \label{fig:qualitative1}
\end{figure}

We provide visual insights on how downsampling affects the tasks by showing qualitative results in Figure~\ref{fig:qualitative1}.
We highlight the contour preservation of the Poisson-based ePDF (\algpoi) estimation in contrast to the Deterministic or Uniform approaches.

\section{Limitations and future work}
While our proposed event PDF estimation and further downsampling method is generic and can be successfully applied in very diverse tasks, we find that generically determine what makes an event \textit{relevant} is an ill-posed problem, as each event has its own relevance depending on the downstream task.
An example we found in our research is based on the Gen1 dataset, where the relevant events to solve that concrete task (car and pedestrian detection) would not be the same as the events needed to solve another task (e.g. road lines segmentation).

Another limitation is on the computation time, as the two baselines, Deterministic and Uniform, are much simpler, thus faster, than the proposed PDF estimators. 
This limitation can be addressed with parallelization and implementation on more appropriate coding languages, and will be considered as future work.
\section{Conclusion}
We introduced \alg, a probabilistic framework for event downsampling that models the process as an online stochastic sampling problem.
By estimating an event importance probability density function (ePDF) on the fly, \alg\ enables adaptive downsampling that prioritizes the most relevant events.
As a specific instance, we proposed a Poisson-based ePDF designed to preserve edge information and compared it against deterministic and uniform baselines.

To evaluate its effectiveness, we tested our method on four diverse tasks: classification, frame interpolation, surface normal estimation, and object detection.
Our experiments, conducted in both zero-shot and fine-tuned settings, demonstrate that structured downsampling strategies significantly outperform naive approaches, particularly in low-event regimes.

Future work includes exploring alternative ePDF formulations, integrating learned event-importance models, and optimizing implementations for real-time deployment on neuromorphic hardware.

{
    \small
    \bibliographystyle{ieeenat_fullname}
    \bibliography{main}

\begin{thebibliography}{17}
\providecommand{\natexlab}[1]{#1}
\providecommand{\url}[1]{\texttt{#1}}
\expandafter\ifx\csname urlstyle\endcsname\relax
  \providecommand{\doi}[1]{doi: #1}\else
  \providecommand{\doi}{doi: \begingroup \urlstyle{rm}\Url}\fi

\bibitem[Araghi et~al.(2024)Araghi, van Gemert, and Tomen]{araghi2024pushing}
Hesam Araghi, Jan van Gemert, and Nergis Tomen.
\newblock Pushing the boundaries of event subsampling in event-based video classification using cnns.
\newblock \emph{arXiv preprint arXiv:2409.08953}, 2024.

\bibitem[Barrios-Avil{\'e}s et~al.(2018)Barrios-Avil{\'e}s, Rosado-Mu{\~n}oz, Medus, Bataller-Mompe{\'a}n, and Guerrero-Mart{\'\i}nez]{barrios2018less}
Juan Barrios-Avil{\'e}s, Alfredo Rosado-Mu{\~n}oz, Leandro~D Medus, Manuel Bataller-Mompe{\'a}n, and Juan~F Guerrero-Mart{\'\i}nez.
\newblock Less data same information for event-based sensors: A bioinspired filtering and data reduction algorithm.
\newblock \emph{Sensors}, 18\penalty0 (12):\penalty0 4122, 2018.

\bibitem[Bi et~al.(2019)Bi, Chadha, Abbas, Bourtsoulatze, and Andreopoulos]{bi2019graph}
Yin Bi, Aaron Chadha, Alhabib Abbas, Eirina Bourtsoulatze, and Yiannis Andreopoulos.
\newblock Graph-based object classification for neuromorphic vision sensing.
\newblock In \emph{Proceedings of the IEEE/CVF international conference on computer vision}, pages 491--501, 2019.

\bibitem[Bisulco et~al.(2020)Bisulco, Ojeda, Isler, and Lee]{bisulco2020near}
Anthony Bisulco, Fernando~Cladera Ojeda, Volkan Isler, and Daniel~Dongyuel Lee.
\newblock Near-chip dynamic vision filtering for low-bandwidth pedestrian detection.
\newblock In \emph{2020 IEEE Computer Society Annual Symposium on VLSI (ISVLSI)}, pages 234--239. IEEE, 2020.

\bibitem[Cohen et~al.(2018)Cohen, Afshar, Orchard, Tapson, Benosman, and van Schaik]{cohen2018spatial}
Gregory Cohen, Saeed Afshar, Garrick Orchard, Jonathan Tapson, Ryad Benosman, and Andre van Schaik.
\newblock Spatial and temporal downsampling in event-based visual classification.
\newblock \emph{IEEE Transactions on Neural Networks and Learning Systems}, 29\penalty0 (10):\penalty0 5030--5044, 2018.

\bibitem[De~Tournemire et~al.(2020)De~Tournemire, Nitti, Perot, Migliore, and Sironi]{de2020large}
Pierre De~Tournemire, Davide Nitti, Etienne Perot, Davide Migliore, and Amos Sironi.
\newblock A large scale event-based detection dataset for automotive.
\newblock \emph{arXiv preprint arXiv:2001.08499}, 2020.

\bibitem[Gehrig et~al.(2019)Gehrig, Loquercio, Derpanis, and Scaramuzza]{Gehrig_2019_ICCV}
Daniel Gehrig, Antonio Loquercio, Konstantinos~G. Derpanis, and Davide Scaramuzza.
\newblock End-to-end learning of representations for asynchronous event-based data.
\newblock In \emph{Int. Conf. Comput. Vis. (ICCV)}, 2019.

\bibitem[Gehrig and Scaramuzza(2023)]{gehrig2023recurrent}
Mathias Gehrig and Davide Scaramuzza.
\newblock Recurrent vision transformers for object detection with event cameras.
\newblock In \emph{Proceedings of the IEEE/CVF Conference on Computer Vision and Pattern Recognition}, pages 13884--13893, 2023.

\bibitem[Ghosh et~al.()Ghosh, Nowotny, and Knight]{ghoshevdownsampling}
Anindya Ghosh, Thomas Nowotny, and James Knight.
\newblock Evdownsampling: a robust method for downsampling event camera data.

\bibitem[Ghosh et~al.(2023)Ghosh, Nowotny, and Knight]{ghosh2023insect}
Anindya Ghosh, Thomas Nowotny, and James~C Knight.
\newblock Insect-inspired spatio-temporal downsampling of event-based input.
\newblock In \emph{Proceedings of the 2023 International Conference on Neuromorphic Systems}, pages 1--5, 2023.

\bibitem[Gruel et~al.(2023{\natexlab{a}})Gruel, Carreras, Garc{\'\i}a, Kupczyk, and Martinet]{gruel2023frugal}
Am{\'e}lie Gruel, Luc{\'\i}a~Trillo Carreras, Marina~Bueno Garc{\'\i}a, Ewa Kupczyk, and Jean Martinet.
\newblock Frugal event data: how small is too small? a human performance assessment with shrinking data.
\newblock In \emph{Proceedings of the IEEE/CVF Conference on Computer Vision and Pattern Recognition}, pages 4092--4099, 2023{\natexlab{a}}.

\bibitem[Gruel et~al.(2023{\natexlab{b}})Gruel, Martinet, Linares-Barranco, and Serrano-Gotarredona]{Gruel_2023_WACV}
Am\'elie Gruel, Jean Martinet, Bernab\'e Linares-Barranco, and Teresa Serrano-Gotarredona.
\newblock Performance comparison of dvs data spatial downscaling methods using spiking neural networks.
\newblock In \emph{Proceedings of the IEEE/CVF Winter Conference on Applications of Computer Vision (WACV)}, pages 6494--6502, 2023{\natexlab{b}}.

\bibitem[Gupta and Bhattacharya(2020)]{gupta2020implementing}
Shriya~TP Gupta and Basabdatta~Sen Bhattacharya.
\newblock Implementing a foveal-pit inspired filter in a spiking convolutional neural network: a preliminary study.
\newblock In \emph{2020 International Joint Conference on Neural Networks (IJCNN)}, pages 1--8. IEEE, 2020.

\bibitem[Muglikar et~al.(2023)Muglikar, Bauersfeld, Moeys, and Scaramuzza]{Muglikar23CVPR}
Manasi Muglikar, Leonard Bauersfeld, Diederik Moeys, and Davide Scaramuzza.
\newblock Event-based shape from polarization.
\newblock In \emph{IEEE / CVF Computer Vision and Pattern Recognition Conference (CVPR)}, 2023.

\bibitem[Orchard et~al.(2015)Orchard, Jayawant, Cohen, and Thakor]{orchard2015converting}
Garrick Orchard, Ajinkya Jayawant, Gregory~K Cohen, and Nitish Thakor.
\newblock Converting static image datasets to spiking neuromorphic datasets using saccades.
\newblock \emph{Frontiers in neuroscience}, 9:\penalty0 437, 2015.

\bibitem[Rizzo et~al.(2023)Rizzo, Schuman, and Plank]{rizzo2023neuromorphic}
Charles~P Rizzo, Catherine~D Schuman, and James~S Plank.
\newblock Neuromorphic downsampling of event-based camera output.
\newblock In \emph{Proceedings of the 2023 Annual Neuro-Inspired Computational Elements Conference}, pages 26--34, 2023.

\bibitem[Tulyakov et~al.(2021)Tulyakov, Gehrig, Georgoulis, Erbach, Gehrig, Li, and Scaramuzza]{tulyakov2021time}
Stepan Tulyakov, Daniel Gehrig, Stamatios Georgoulis, Julius Erbach, Mathias Gehrig, Yuanyou Li, and Davide Scaramuzza.
\newblock Time lens: Event-based video frame interpolation.
\newblock In \emph{Proceedings of the IEEE/CVF conference on computer vision and pattern recognition}, pages 16155--16164, 2021.

\end{thebibliography}
}


\end{document}